\documentclass{article}
\usepackage{iclr2026_conference,times}
\iclrfinalcopy
\usepackage[T1]{fontenc}
\usepackage{amsmath,amssymb}
\usepackage{booktabs,tabularx}
\usepackage{xcolor}
\usepackage{graphicx}
\usepackage{needspace}
\usepackage{tikz}
\usetikzlibrary{arrows.meta,positioning,calc}
\usepackage[hidelinks]{hyperref}
\hypersetup{
  pdftitle={FastDC: Efficient Execution of Hallucination Detection and Correction for Large Language Models},
  pdfauthor={Jiangneng Li, Xianzhi Zeng, Peiqi Yin, Wenchao Zhou, Gao Cong, Cheng Long}
}
\usepackage{url}
\newcommand{\workname}{FastDC}

\title{\workname{}: Efficient Execution of\\Hallucination Detection and Correction\\for Large Language Models}
\author{%
  \begin{minipage}[t]{0.31\textwidth}\raggedright
    \textbf{Jiangneng Li}\\
    Nanyang Technological\\
    University\\
    {\fontsize{8}{10}\selectfont\href{mailto:jiangnen002@e.ntu.edu.sg}{\texttt{jiangnen002@e.ntu.edu.sg}}}
  \end{minipage}\hfill
  \begin{minipage}[t]{0.31\textwidth}\raggedright
    \textbf{Xianzhi Zeng}\\
    Nanyang Technological\\
    University\\
    {\fontsize{8}{10}\selectfont\href{mailto:xianzhi001@e.ntu.edu.sg}{\texttt{xianzhi001@e.ntu.edu.sg}}}
  \end{minipage}\hfill
  \begin{minipage}[t]{0.33\textwidth}\raggedright
    \textbf{Peiqi Yin}\\
    The Chinese University\\
    of Hong Kong\\
    {\fontsize{8}{10}\selectfont\href{mailto:pqyin22@cse.cuhk.edu.hk}{\texttt{pqyin22@cse.cuhk.edu.hk}}}
  \end{minipage}\\[10pt]
  \begin{minipage}[t]{0.33\textwidth}\raggedright
    \textbf{Wenchao Zhou}\\
    Alibaba Cloud Computing\\
    {\fontsize{8}{10}\selectfont\href{mailto:zwc231487@alibaba-inc.com}{\texttt{zwc231487@alibaba-inc.com}}}
  \end{minipage}\hfill
  \begin{minipage}[t]{0.33\textwidth}\raggedright
    \textbf{Gao Cong}\\
    Nanyang Technological University\\
    {\fontsize{8}{10}\selectfont\href{mailto:gaocong@ntu.edu.sg}{\texttt{gaocong@ntu.edu.sg}}}
  \end{minipage}\hfill
  \begin{minipage}[t]{0.33\textwidth}\raggedright
    \textbf{Cheng Long}\\
    Nanyang Technological University\\
    {\fontsize{8}{10}\selectfont\href{mailto:c.long@ntu.edu.sg}{\texttt{c.long@ntu.edu.sg}}}
  \end{minipage}}
\makeatletter
\renewcommand{\@maketitle}{%
  \vbox{\hsize\textwidth
    {\LARGE\scshape\raggedright\@title\par}
    \vskip 0.15in
    {\normalsize\raggedright\@author\par}
    \vskip 0.3in}}
\makeatother
\begin{document}
\maketitle
\suppressfloats[t]
\lhead{Preprint}
\begin{abstract}
Detecting hallucinations and other errors in LLM-generated responses is essential for accuracy-critical applications, such as financial analysis and healthcare.
Sampling-based hallucination detection and evidence-based correction turn a
single answer into a pipeline of interdependent language-model calls.
However, the evaluation and verification required by these workflows
increase the cost and latency of producing a final answer, motivating more
efficient execution.
This pipeline contains two useful forms of redundancy: independently sampled
responses can share identical token prefixes, while corrections often retain
portions of the original answer. We present \workname{}, an inference serving
system for
efficient execution of hallucination detection and correction.
It shares model computation across identical prefixes during detection
and query generation, and verifies existing text as speculative candidates
during correction. \workname{} overlaps correction checks while preserving
revision order and separates attention execution to reduce interference
with detection. Experiments with Qwen2.5-14B 
demonstrate $1.55\times$ and $1.22\times$ throughput speedups for short
and long answers, respectively. Mean job completion time falls by
34.4\% and 18.4\%, and GPU execution time by 36.2\% and 25.7\%, respectively.
\end{abstract}
\section{Introduction}
\label{sec:intro}

Hallucination detection and correction are increasingly important as LLMs
enter high-stakes applications, such as financial analysis and healthcare. For instance, Azure's groundedness API checks generated
text against source material and offers an optional correction feature
\citep{microsoft2026groundedness}. In finance, BNY describes using independent
model analyses to validate financial statements, with disagreements escalated
for human review \citep{bny2026fundoperations}. Such checks place additional
work on the path to a usable result, making their execution efficiency part
of the serving problem.

\begin{figure}[t]
\centering
{%
\begin{tikzpicture}[x=1cm,y=1cm,font=\sffamily\fontsize{8}{9.5}\selectfont,
 box/.style={draw=black!45,rounded corners=2pt,align=center,
   minimum height=1.20cm,inner xsep=4pt,inner ysep=3pt},
 arr/.style={-{Stealth[length=1.5mm]},thick,draw=black!65},
 route/.style={font=\sffamily\fontsize{7.5}{9}\selectfont,align=center}]
\useasboundingbox (0,0.05) rectangle (13.7,2.15);
\node[box,fill=black!3,minimum width=2.1cm] (a) at (1.1,0.80)
 {\textbf{Question}\\When did the\\museum open?};
\node[box,fill=blue!5,minimum width=3.8cm,right=0.5cm of a] (d)
 {\textbf{Generation \& detection}\\
  Answer: \textbf{1997}\\
  Samples: 1997, 1998, $\ldots$};
\node[box,fill=green!7,minimum width=3.2cm,right=1.5cm of d] (c)
 {\textbf{Correction}\\
  Evidence: \textbf{1998}\\
  Revise: $1997 \rightarrow 1998$};
\node[box,fill=black!3,minimum width=1.75cm,right=0.5cm of c] (f)
 {Final\\answer};
\draw[arr] (a) -- (d);
\draw[arr] (d) -- node[route,above]{High\\uncertainty} (c);
\draw[arr] (c) -- (f);
\draw[arr] (d.north) -- ($(d.north)+(0,0.25)$)
 -- node[route,above]{Low uncertainty}
 ($(f.north)+(0,0.25)$) -- (f.north);
\end{tikzpicture}}
\caption{{Illustrative workflow on a shared LLM serving engine.
The initial answer and auxiliary responses are generated concurrently;
answers with high uncertainty proceed to evidence-based correction.}}
\label{fig:overview}
\end{figure}
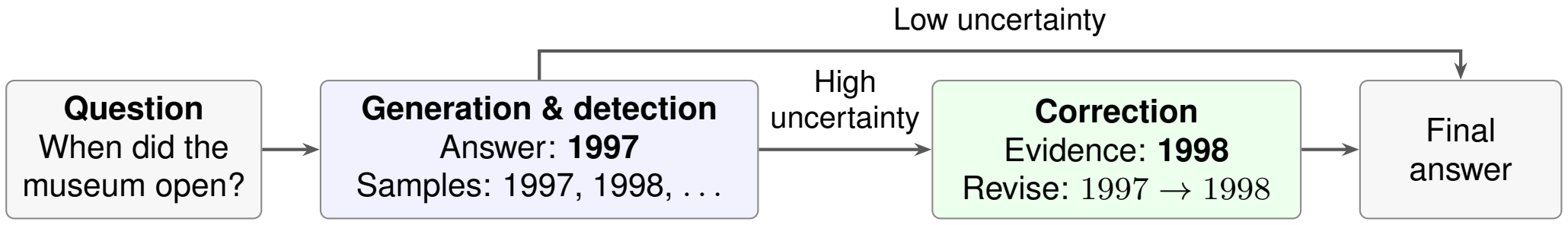

We study a detection--correction workflow~\citep{guo2026finground} in which detection identifies
potentially unreliable answers and correction verifies and revises flagged
answers using external evidence. {Figure~\ref{fig:overview} summarizes this workflow.} Sampling-based detectors such as semantic entropy,
SelfCheckGPT, and confidence--consistency methods provide useful evidence
about unreliable outputs
{\citep{kuhn2023semantic,manakul2023selfcheck,vashurin2025cocoa,pawar2026tokens}}.
Research-and-revision pipelines such as RARR use external evidence to check
and revise generated text \citep{gao2023rarr}. Both detection and correction
involve LLM inference, and their requests can execute on a shared serving
engine such as vLLM \citep{kwon2023pagedattention} or SGLang
\citep{zheng2024sglang}. In this setting, we optimize the complete pipeline and measure
its cost and job completion time.

In this workflow, we identify two forms of computational redundancy and exploit them to improve execution efficiency. Specifically, independent samples
can have identical complete token histories; their model forward can be
shared before each logical sample draws independently. 
Correction has a different opportunity: much of a revised answer is already
present in the previous wrong answer or retrieved evidence. That text can provide
speculative candidates, provided every accepted token is verified under the
correct current prompt.

Reducing job completion time is critical for the timely delivery of final answers. However, optimizing detection and correction independently does not necessarily achieve this goal.
This is because when the two phases share an inference engine, accelerating
one phase can slow the other through resource interference, ultimately
increasing mean job completion time.
Detection decodes one token per step, whereas speculative correction verifies several; mixing them in one attention call forces detection off its fast single-token path.
This
motivates partial decoupling at the attention boundary, while retaining joint
batching for the linear layers.

In this paper, we present \workname{}, a 
serving
system for hallucination detection
and correction built on vLLM \citep{kwon2023pagedattention}.
{Our design reuses language model
computation across samples and text available within the same workflow,
overlaps correction checks while preserving revision order, and reduces
interference between 
detection and correction
in the shared engine.}

{For detection and query generation, we share forward computation
across independent samples of the same prompt whenever their generated
prefixes are identical.}
{A shared state with multiplicity $m$ represents $m$ independent
samples with the same prompt and generated prefix. We compute the next-token
distribution once and draw a token independently for each sample; samples
that diverge resume separate computation.} This reduces repeated model forwards while
preserving independent sampling and response multiplicity.
{At the workflow level, we stop auxiliary sampling early once
completed responses guarantee that the remaining samples cannot change the
detector's decision under the full sampling budget.} 
{For correction, we accelerate generation through speculative
decoding, using text already present in the current prompt or generated
output as candidate continuations.}
{Native n-gram lookup proposes continuations from the current
prompt and generated history, reusing available text such as the current
answer and retrieved evidence.}
{For example, the previously generated answer evaluated during
hallucination detection can supply candidate tokens when included in the
correction prompt. The LLM checks these reused tokens against its own
predictions under that prompt and accepts them up to the first mismatch.
This allows several tokens to be produced in a single forward pass.}
{Furthermore, correction may require multiple checks on different
parts of the generated answer. \workname{} allows multiple checks to run in
parallel while applying their results in the original order. If an earlier
correction changes the answer, later checks based on the previous version
are discarded and rerun on the updated answer.}

We coordinate shared sampling and speculative correction within the
same engine by separating attention according to physical query length.
Linear layers process both phases together,
while single-query decode and multi-query verification use separate
attention paths. This retains the batching benefits of joint matrix
multiplications while mitigating the attention interference between the two
phases. This partitioning also accommodates prefill alongside
decoding and selects a suitable attention path for the single-token group.

We summarize our contributions as follows:\nopagebreak[4]
\begin{itemize}
  \item {\textbf{A unified execution system for detection and correction.}
  We develop \workname{}, a unified execution system built on vLLM that
  accelerates the complete hallucination detection--correction workflow.
  For detection and query generation, it shares model computation across
  independent samples. For correction, it reuses existing text as speculative
  candidates that the LLM verifies before acceptance. The system further
  reduces unnecessary auxiliary generation and overlaps correction checks
  while preserving the detector's fixed-budget decision and the order of
  answer revisions.}
  \item \textbf{Efficient GPU execution for shared sampling and
  speculative correction.}
  We decouple attention execution for single-token decoding and multi-token
  verification to reduce interference, while retaining joint batching for
  matrix multiplications. The execution paths accommodate mixed prefill
  and decoding as well as smaller groups of single-token requests.
  \item \textbf{End-to-end and component-level evaluation.}
  Experiments with Qwen2.5-14B demonstrate $1.55\times$ and
  $1.22\times$ throughput speedups for short and long answers, respectively,
  over the native vLLM workflow.
  We find that speculative correction alone reduces long-answer throughput by $16\%$ due to attention interference, and that attention decoupling reverses this loss with a $36\%$ gain.
\end{itemize}

\section{Preliminaries}
\label{sec:contract}

\subsection{{Detection--correction workflow}}

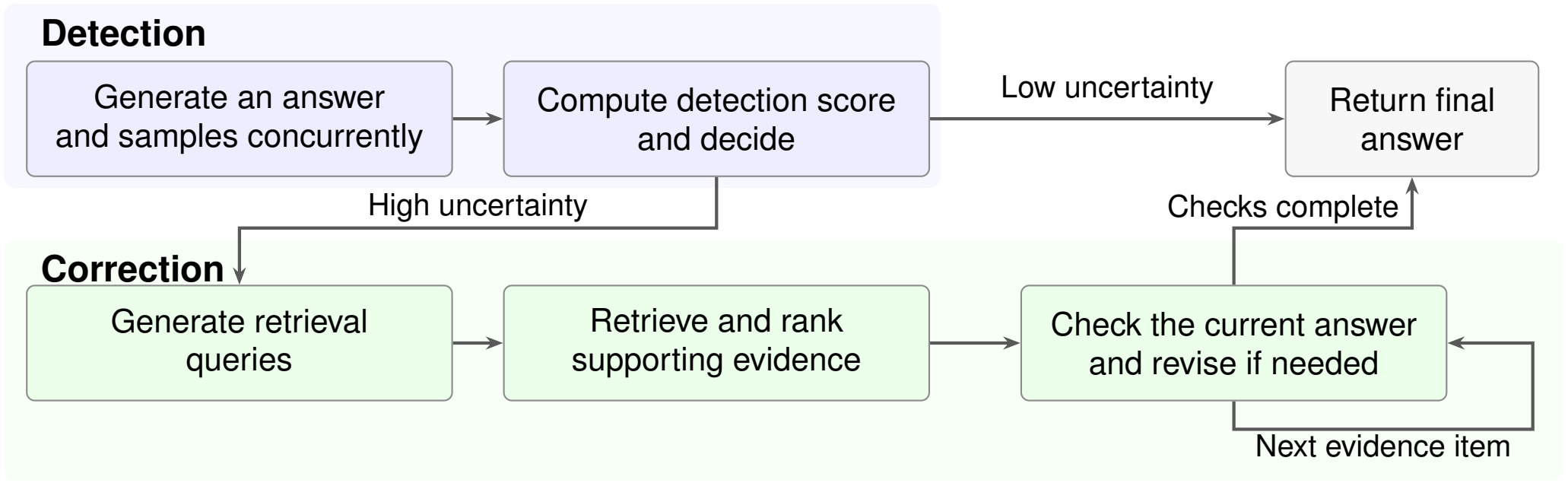
\begin{figure}[htbp]
\centering
{%
\begin{tikzpicture}[x=1cm,y=1cm,font=\sffamily\fontsize{8}{9.5}\selectfont,
 box/.style={draw=black!45,rounded corners=2pt,align=center,
   minimum width=3.7cm,minimum height=1.0cm,
   inner xsep=4pt,inner ysep=3pt},
 arr/.style={-{Stealth[length=1.5mm]},thick,draw=black!65},
 route/.style={font=\sffamily\fontsize{7.5}{9}\selectfont,align=center},
 stage/.style={font=\sffamily\fontsize{9}{10.5}\selectfont\bfseries,
   anchor=west}]
\useasboundingbox (0,0.30) rectangle (13.7,4.55);
\fill[blue!3,rounded corners=3pt] (0.05,2.90) rectangle (8.2,4.50);
\fill[green!3,rounded corners=3pt] (0.05,0.35) rectangle (13.6,2.45);
\node[stage] at (0.25,4.25) {Detection};
\node[stage] at (0.25,2.2) {Correction};

\node[box,fill=blue!7] (a) at (2.1,3.50)
 {Generate an answer\\and samples concurrently};
\node[box,fill=blue!7] (d) at (6.25,3.50)
 {Compute detection score\\and decide};
\node[box,fill=black!3,minimum width=2.2cm] (f) at (12.3,3.50)
 {Return final\\answer};
\node[box,fill=green!8] (q) at (2.1,1.55)
 {Generate retrieval\\queries};
\node[box,fill=green!8] (r) at (6.25,1.55)
 {Retrieve and rank\\supporting evidence};
\node[box,fill=green!8] (c) at (10.75,1.55)
 {Check the current answer\\and revise if needed};

\draw[arr] (a) -- (d);
\draw[arr] (d) -- node[route,above]{Low uncertainty} (f);
\draw[arr] (d.south) -- (6.25,2.55)
 -- node[route,above,inner ysep=1pt]{High uncertainty}
 (2.1,2.55) -- (q.north);
\draw[arr] (q) -- (r);
\draw[arr] (r) -- (c);
\draw[arr] (c.north) -- (10.75,2.55)
 -- (12.3,2.55) -- (f.south);
\node[route,anchor=south east,inner xsep=3pt,inner ysep=1pt]
 at (12.3,2.55) {Checks complete};
\draw[arr] (c.south) -- (10.75,0.80)
 -- node[route,below,inner ysep=1pt]{Next evidence item} (13.35,0.80)
 -- (13.35,1.55) -- (c.east);
\end{tikzpicture}}
\caption{{The reference answering workflow.
Detection either returns the original answer or triggers
retrieval and ordered correction. Each successive evidence check uses the
latest answer, including any preceding revisions.}}
\label{fig:workflow}
\end{figure}

{\paragraph{Generation and detection.}
A workflow starts with a user question. The LLM generates the initial answer
concurrently with additional independently sampled answers for the same
question, which we call auxiliary responses~\citep{pawar2026tokens, manakul2023selfcheck}. The detector evaluates the
completed initial answer using these responses, computes a detection score,
and compares it with a threshold. Answers with low uncertainty are returned
directly; those with high uncertainty proceed to evidence-based correction.
Figure~\ref{fig:workflow} shows the complete workflow.}

{\paragraph{Evidence-based correction.}
For an answer with high uncertainty, the LLM generates queries to ask for external
evidence, and then correct the answer. The
workflow deduplicates and orders these queries, retrieves relevant passages,
and ranks the evidence. It then checks the current answer against each
evidence item, revising the answer when needed~\citep{guo2026finground}. We call the LLM invocation
that decides whether a revision is needed a \emph{gate}, and the invocation
that produces the revised answer an \emph{editor}. Checks and revisions
follow a fixed order, so each check reads the answer after preceding
revisions. The final answer is returned after all checks.}

{\paragraph{Workflow formulation.}
For question $x$, let $y$ denote the original answer and $z_{1:N}$ the $N$
independent auxiliary responses. The fixed-budget detector computes}
\begin{equation}
 b=\mathbf{1}\{f_N(x,y,z_{1:N})>\tau\},
 \label{eq:decision}
\end{equation}
{with a fixed threshold $\tau$. If $b=0$, the service returns $y$;
if $b=1$, it starts correction. Let $y^{(0)}=y$. At correction step $j$, the
gate reads the current committed answer $y^{(j-1)}$ and fixed evidence $e_j$.
An edit, when required, produces $y^{(j)}$; otherwise
$y^{(j)}=y^{(j-1)}$.}
\paragraph{Execution setting.}
All LLM stages share one model and GPU in this study. Semantic encoding,
retrieval, and reranking execute in CPU workers. Different workflows may
occupy different stages concurrently. 

\subsection{{Optimization objective}}

A generated answer must pass through detection and any required correction before it is delivered. Job completion time therefore determines how long users or downstream applications must wait for the final answer.
{In this paper, we aim to reduce GPU execution time and job completion
time and increase final-answer throughput while preserving the results
of the detection--correction workflow, including its detection decisions and
final answers.}

{\paragraph{Job completion time (JCT).}
We treat the detection--correction workflow for each question as a job.
For question $x$, let $t_{\mathrm{arr}}(x)$ denote its arrival time at the
service and $t_{\mathrm{final}}(x)$ the time its complete final answer is
delivered. We define its JCT as
\begin{equation}
 T_{\mathrm{JCT}}(x)=t_{\mathrm{final}}(x)-t_{\mathrm{arr}}(x).
 \label{eq:e2e-latency}
\end{equation}
This wall-clock interval includes admission queueing, computation, and
waiting between workflow stages.}

\paragraph{Throughput.}
For a measurement window $W=(t_a,t_b]$, let $N_{\mathrm{delivered}}(W)$ count
the final answers successfully delivered within $W$. We define throughput as
\begin{equation}
 R_{\mathrm{final}}(W)=\frac{N_{\mathrm{delivered}}(W)}{t_b-t_a},
 \label{eq:final-throughput}
\end{equation}
measured in answers per second. 

{\paragraph{Independent sampling.}
For a fixed sampling budget, e.g., 10 samplings, each auxiliary response must remain an
independent draw from the original token generation distribution. Identical
responses still count as separate samples when computing the hallucination detection
score. Execution optimizations may share computation, but must preserve
these sampling and counting semantics.}

{\paragraph{Ordered correction.}
We aim to optimize the correction process described above while preserving its outputs. Therefore, to retain the output, we keep the search queries (natural-language questions used to retrieve evidence), retrieved evidence, LLM prompts, and generation settings unchanged. Further, \workname{} has to preserve the original order of checks and revisions because an earlier revision can change the answer that later checks need to evaluate.
\section{Joint Execution Design}
\label{sec:method}

Figure~\ref{fig:reuse} illustrates three key mechanisms through concrete
examples: shared sampling, verified text reuse during correction, and
concurrent checks that preserve revision order.

\begin{figure}[t]
\centering
\scalebox{0.9}{%
\begin{tikzpicture}[x=1cm,y=1cm,
 font=\sffamily\fontsize{8}{9.5}\selectfont,
 box/.style={draw=black!40,rounded corners=2pt,align=center,
   inner xsep=4pt,inner ysep=4pt},
 token/.style={box,minimum height=.55cm,inner xsep=3pt,inner ysep=2pt},
 sampletoken/.style={token,inner xsep=1pt},
 arr/.style={-{Stealth[length=1.4mm]},thick,draw=black!65},
 paneltitle/.style={font=\sffamily\fontsize{9}{10.5}\selectfont\bfseries},
 boxtitle/.style={font=\sffamily\fontsize{8}{9.5}\selectfont\bfseries},
 small/.style={font=\sffamily\fontsize{7}{8.5}\selectfont},
 labeltext/.style={small,anchor=west,text=black!70}]
\useasboundingbox (0,.25) rectangle (13.7,11.4);
\begin{scope}[yshift=3.6cm]
\fill[blue!3,rounded corners=3pt] (.05,.1) rectangle (6.65,7.75);
\fill[green!3,rounded corners=3pt] (7.05,.1) rectangle (13.65,7.75);
\node[paneltitle] at (3.35,7.4)
 {(a) Shared sampling};
\node[small,text=black!65] at (3.35,7.0)
 {Detection and repeated query generation};
\node[paneltitle] at (10.35,7.4)
 {(b) Target-verified correction};
\node[small,text=black!65] at (10.35,7.0)
 {Editor example: change the opening year};

\node[box,fill=white,text width=5.6cm,minimum height=1.05cm] (history)
 at (3.35,6.1)
 {Question: What is the capital of France?\\
  Same history $h$: \texttt{The capital is}\\
  \textbf{4 logical samples; multiplicity $m=4$}};
\node[box,fill=blue!12,text width=5.6cm,minimum height=.8cm] (forward)
 at (3.35,4.85)
 {\textbf{One model forward at $h$}\\
  One shared distribution $p(\text{next token}\mid h)$};
\draw[arr] (history) -- (forward);
\node[box,fill=white,minimum width=5.9cm,minimum height=1.15cm] (draws)
 at (3.35,3.55) {};
\node[small] at (3.35,3.88)
 {Four independent draws, one per logical sample};
\node[sampletoken,fill=blue!10,minimum width=1.15cm] (s1) at (1.15,3.35)
 {$s_1$: \texttt{Paris}};
\node[sampletoken,fill=blue!10,minimum width=1.15cm] (s2) at (2.6,3.35)
 {$s_2$: \texttt{Paris}};
\node[sampletoken,fill=blue!10,minimum width=1.15cm] (s3) at (4.05,3.35)
 {$s_3$: \texttt{Paris}};
\node[sampletoken,fill=orange!15,minimum width=1.15cm] (s4) at (5.5,3.35)
 {$s_4$: \texttt{Lyon}};
\path let \p1=($(s2.west)-(s1.east)$) in
 \pgfextra{\xdef\samplegap{\x1}};
\draw[arr] (forward) -- (draws.north);
\node[box,fill=blue!10,text width=2.55cm,minimum height=.85cm] (paris)
 at (1.9,1.95) {$h+\texttt{Paris}$\\Multiplicity $m=3$};
\node[box,fill=orange!12,text width=2.55cm,minimum height=.85cm] (lyon)
 at (4.8,1.95) {$h+\texttt{Lyon}$\\Multiplicity $m=1$};
\draw[arr] (s1.south) -- ([xshift=-.55cm]paris.north);
\draw[arr] (s2.south) -- (paris.north);
\draw[arr] (s3.south) -- ([xshift=.55cm]paris.north);
\draw[arr] (s4.south) -- (lyon.north);
\node[box,fill=white,text width=2.55cm,minimum height=.85cm] (nextp)
 at (1.9,.8) {One forward\\3 IID draws};
\node[box,fill=white,text width=2.55cm,minimum height=.85cm] (nextl)
 at (4.8,.8) {One forward\\1 IID draw};
\draw[arr] (paris) -- (nextp);
\draw[arr] (lyon) -- (nextl);

\node[box,fill=white,text width=5.6cm,align=left,minimum height=1.25cm] (context)
 at (10.35,6.1)
 {\textbf{Original answer:} The museum first opened in
  \textcolor{red!70!black}{1997} and is located in central Rome.\\[2pt]
  \textbf{Evidence:} Opening year: \textcolor{blue!70!black}{1998}.};
\node[small,align=left,text width=5.8cm] at (10.35,5.15)
 {Editor prefix: \texttt{The museum first opened}};

\node[box,fill=white,minimum width=5.9cm,minimum height=2.4cm] (verify)
 at (10.35,3.55) {};
\node[boxtitle] at (10.35,4.4)
 {Target verification: one forward};
\node[labeltext,inner xsep=0pt] at (7.55,3.85) {N-gram};
\node[labeltext,inner xsep=0pt] at (7.55,3.05) {Editor};
\node[token,fill=green!15,minimum width=.6cm] (draftin) at (8.8,3.85) {\texttt{in}};
\node[token,fill=red!12,minimum width=.8cm,inner xsep=1pt,right=\samplegap of draftin] (draftyear) {\texttt{1997}};
\node[token,fill=black!5,text=black!55,inner xsep=1pt,right=\samplegap of draftyear] (draftrest)
 {\texttt{and is located in}};
\node[token,fill=green!15,minimum width=.6cm] (predin) at (draftin.center |- 0,3.05) {\texttt{in}};
\node[token,fill=blue!12,minimum width=.8cm,inner xsep=1pt] (predyear) at (draftyear.center |- 0,3.05) {\texttt{1998}};
\node[small,text=black!55,align=center] at (draftrest.center |- 0,3.05)
 {Previous suffix};
\draw[arr] (draftin.south) -- (predin.north);
\draw[arr] (draftyear.south) -- (predyear.north);
\node[small,text=green!45!black] at (predin.center |- 0,2.55) {Accept};
\node[small,text=blue!70!black,anchor=west,inner xsep=0pt,xshift=.12cm] at (predyear.west |- 0,2.55) {Output \texttt{1998}};

\node[box,fill=white,minimum width=5.9cm,minimum height=1.8cm] (resume)
 at (10.35,1.25) {};
\node[boxtitle]
 at (10.35,1.87) {Resume decoding; match a new suffix};
\node at (10.35,1.38)
 {\texttt{\ldots\ \textcolor{blue!70!black}{1998} \underline{and is located in}}};
\node[token,fill=green!15,minimum width=2cm] (continuation)
 at (8.95,.72) {\texttt{central Rome.}};
\node[small] (commit) at (11.65,.72) {Verify, then commit};
\draw[arr] (continuation.east) -- (commit.west);
\end{scope}
\fill[blue!2,rounded corners=3pt] (.05,.3) rectangle (13.65,3.4);
\node[paneltitle] at (6.85,3.12)
 {(c) Gate lookahead with ordered answer revisions};
\node[small,text=black!65] at (6.85,2.73)
 {Same museum example: the editor revises the opening year from
  \textcolor{red!70!black}{1997} to \textcolor{blue!70!black}{1998}.};
\draw[arr,draw=black!35] (2.13,2.4) -- (13.3,2.4);
\node[small,anchor=south east,text=black!60] at (13.3,2.43) {Time};

\node[anchor=west,align=left] at (.2,1.88) {Current gate\\and editor};
\node[anchor=west] at (.2,1.0) {Next gate};

\node[box,fill=blue!10,minimum width=2.55cm,minimum height=.63cm,
 inner ysep=2pt] (cg1) at (3.405,1.88)
 {Gate$_i(A_0)$\\Year: disagree};
\node[box,fill=green!12,minimum width=2.7cm,minimum height=.63cm,
 inner ysep=2pt] (cedit) at (6.31,1.88)
 {Editor$_i$\\\textcolor{red!70!black}{1997} $\rightarrow$
  \textcolor{blue!70!black}{1998}};
\draw[arr] (cg1.east) -- (cedit.west);
\draw[arr] (cedit.east) -- (8.17,1.88);
\fill[green!45!black] (8.17,1.88) circle (1.6pt);
\node[small,anchor=south,text=green!45!black] at (8.17,2.06)
 {Commit $A_1$};
\node[small,anchor=west,text=red!70!black] at (8.5,1.88)
 {Answer changed};

\node[box,fill=blue!10,minimum width=4.22cm,minimum height=.63cm,
 inner ysep=2pt] (cold) at (4.24,1.0)
 {Gate$_{i+1}(A_0)$\\Check location on $A_0$};
\draw[draw=black!45,dashed,thick] (cold.east) -- (8.17,1.0);
\draw[draw=red!70!black,dashed,thick] (8.17,1.8) -- (8.17,.78);
\draw[draw=red!70!black,line width=1pt] (8.07,.9) -- (8.27,1.1);
\draw[draw=red!70!black,line width=1pt] (8.07,1.1) -- (8.27,.9);
\node[small,text=black!60] at (7.18,1.23) {Result waits};
\node[small,text=red!70!black] at (7.75,.52) {Discard old result};

\node[box,fill=blue!10,minimum width=3.78cm,minimum height=.63cm,
 inner ysep=2pt] (cnew) at (10.71,1.0)
 {Gate$_{i+1}(A_1)$\\Recheck location on $A_1$};
\draw[arr] (8.33,1.0) -- (cnew.west);
\draw[arr] (cnew.east) -- (13.12,1.0);
\node[text=green!45!black,inner sep=0pt] at (13.18,1.0) {$\checkmark$};
\node[small,text=blue!70!black] at (10.71,.52)
 {Use the new result in order};
\end{tikzpicture}
}
\caption{{Key mechanisms of \workname{}.
(a) Samples with identical prompts and generated prefixes share computation
while retaining independent draws and multiplicity.
(b) Under its current prompt, the correction model verifies multiple candidate
tokens from the original answer in one forward, accepting matching tokens
and generating \emph{1998} instead of \emph{1997}.
(c) Checks overlap, but results and revisions are applied in order. The early
location check is reused if the answer is unchanged and rerun otherwise.}}
\label{fig:reuse}
\end{figure}
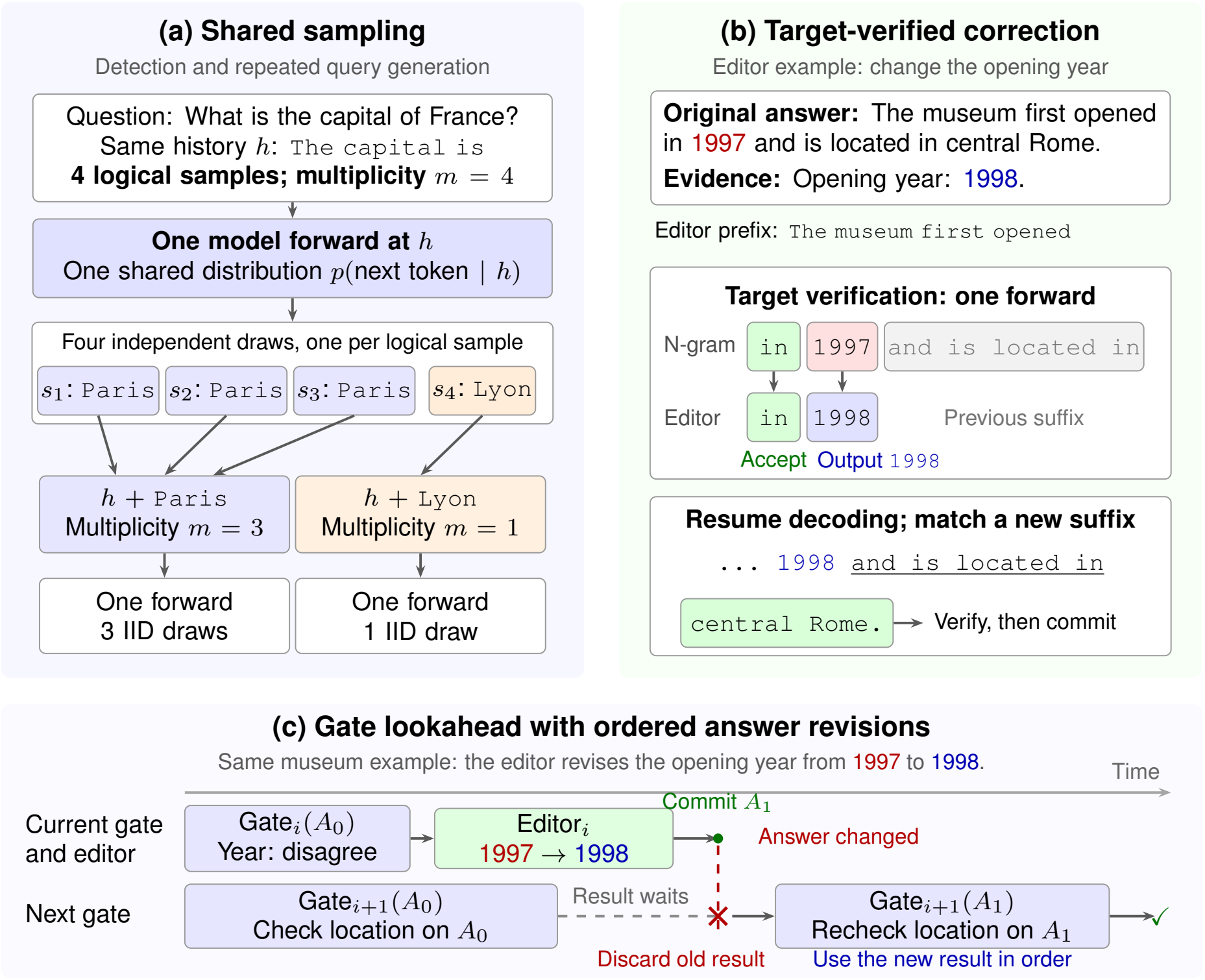

\subsection{Shared states for hallucination detection and query generation}

{For detection and query generation, \workname{} shares model
computation across independent samples that have the same prompt and
generated prefix. Figure~\ref{fig:reuse}a illustrates four samples with the
shared prefix ``The capital is''. Instead of computing the next-token
distribution four times, \workname{} computes it once and draws a token
independently for each sample. In this example, three samples generate
``Paris'' and one generates ``Lyon''. The next decoding step therefore
processes two distinct prefixes rather than four: the three ``Paris'' samples
continue to share computation, while the ``Lyon'' sample proceeds separately.
Each sample retains its own independent draw at every step.}

{Formally, let $h=(w_1,\ldots,w_\ell)$ be the complete token
history, consisting of the prompt followed by the generated prefix. Let
$\mathcal I_h$ contain the logical samples with this history and the same
sampling configuration; its \emph{multiplicity} is $m_h=|\mathcal I_h|$.
Let $\mathcal V$ denote the token vocabulary. The next-token sampling
distribution at history $h$ is defined by
\begin{equation}
 q_h(v)=\Pr(\text{next token}=v\mid h),\qquad v\in\mathcal V.
 \label{eq:shared-distribution}
\end{equation}
It is obtained from the model logits using the same sampling settings and token constraints as separate execution.
A single forward at $h$ supplies these logits for
all samples in $\mathcal I_h$. Each sample draws its next token $T_i$ as
\begin{equation}
 T_i\sim q_h,\qquad i\in\mathcal I_h,\qquad\text{independently}.
 \label{eq:shared-sampling}
\end{equation}
Let $v_1,\ldots,v_K$ be the distinct tokens drawn, and let $\Vert$ denote
token concatenation. The resulting branch histories and multiplicities are
\begin{equation}
 h_j=h\mathbin{\Vert}v_j,\qquad
 m_j=\sum_{i\in\mathcal I_h}\mathbf{1}\{T_i=v_j\},\qquad
 \sum_{j=1}^{K}m_j=m_h,
 \label{eq:shared-branches}
\end{equation}
The multiplicities follow from the independent draws; they are not fixed
in advance. In Figure~\ref{fig:reuse}a,
$m_h=4$ splits into $m_1=3$ and $m_2=1$. The same rule applies at subsequent
steps, sharing computation within each continuing branch. For a fixed
$q_h$, this produces the same joint sampling distribution as separate
execution while retaining every logical sample.}

{Each sample retains its own request state for independent
sampling and completion handling. \workname{} computes shared logits once
and expands them to one row per logical sample before native sampling.
To preserve vLLM's existing KV management, it copies only newly computed
KV entries into the other requests' allocated cache slots.
Sharing requires an identical complete conditioning history.}

{This mechanism complements vLLM's automatic prefix caching.
Prefix caching reuses previously computed prompt blocks, while our method
additionally shares computation among samples that continue to generate
identical prefixes. 
Appendix~\ref{app:shared-implementation} describes the
eligibility checks and integration with asynchronous execution.}

\subsection{Certified Early Stopping}
\label{sec:bounds}

\noindent{We implement early stopping with bounds on the fixed-budget
detection score. Let $o_t$ contain completed responses and available target
statistics, and let $\mathcal C(o_t)$ contain every full outcome consistent
with those observations and the fixed generation budget. A valid interval
satisfies
\begin{equation}
 L_t\leq f_N(c)\leq U_t\quad \text{for all }c\in\mathcal C(o_t).
 \label{eq:certificate}
\end{equation}
The comparison result between $f_N(c)$ and $\tau$ is fixed when $U_t\leq\tau$ or $L_t>\tau$.
Early stopping therefore preserves the decision of the $N$-sample
detector at the same threshold, without requiring all $N$ responses to finish.}

The target and all $N=10$ auxiliaries begin concurrently. For the
length-normalized CoCoA instance \citep{vashurin2025cocoa,pawar2026tokens},
let $A$ be the complete target's mean negative log-probability and let
$d_i=1-\cos(e(y),e(z_i))\in[0,2]$. The score is
\begin{equation}
 f_N^{\mathrm C}=\frac{A}{N}\sum_{i=1}^{N}d_i.
\end{equation}
For any completed subset $S$ of size $m$, with $D_S=\sum_{i\in S}d_i$,
\begin{equation}
 L_S=\frac{AD_S}{N},\qquad
 U_S=\frac{A\{D_S+2(N-m)\}}{N}.
 \label{eq:cocoa-bound}
\end{equation}
The interval follows from the range of each remaining distance and covers
arbitrary future outputs. {For CoCoA, we compute $A$ from the
completed target answer and evaluate $d_i$ only for completed auxiliary
responses. For each unfinished response, we use the minimum and maximum
possible distances, 0 and 2, to compute the score bounds.}

{Once the target answer is complete, \workname{} updates the score
bounds as completed auxiliary responses become available, while the remaining
responses continue generating. Identical responses reuse the same embedding,
but each sample still contributes separately to the score. When the bounds
establish the decision, \workname{} cancels unfinished auxiliary responses
and retains the target answer for return or correction. Implementation
details are given in Appendix~\ref{app:bounds}.}
The same rule applies to any detection score whose final value can be bounded from partial observations.

\subsection{Verified text reuse in correction}
\label{sec:correction}

\noindent{The correction stage follows the
research-and-revision workflow of RARR \citep{gao2023rarr}, using local
retrieval and the same LLM for query generation, checking, and editing.
To gather evidence, the model independently generates three sets of
verification questions about the original answer. We merge duplicate
questions and establish a fixed processing order. For each question, we
retrieve passages and rerank them by relevance before checking whether
the current answer needs revision.}

{Correction can leave much of the original text unchanged.
In Figure~\ref{fig:reuse}b, the original answer gives the museum's opening
year as 1997, while the retrieved evidence gives 1998. When the editor reaches
``The museum first opened'', the original answer supplies the candidate
continuation ``in 1997 and is located in''. Under the correction prompt,
the model accepts ``in'' but generates ``1998'' instead of the proposed
``1997''. Later, matching ``and is located in'' supplies ``central Rome.''
as a new candidate, which the model verifies before accepting. Reusing these
spans can advance several tokens in a single forward pass.}

{We implement this reuse with native n-gram prompt lookup and
greedy speculative decoding \citep{leviathan2023speculative}. Let
$\tilde t_{1:k}=(\tilde t_1,\ldots,\tilde t_k)$ denote a sequence of $k$
candidate tokens retrieved from text already in the prompt or generated
output.  At position $j$,
the model predicts the next token $g_j$ using the current prompt, the
accepted output prefix, and the preceding candidate tokens
$\tilde t_{1:j-1}$. Here, $g_j$ is the highest-probability token allowed by
the call's output constraints. A single causal forward evaluates
these predictions for $j=1,\ldots,k+1$. We accept the
longest matching candidate prefix, whose length is
\begin{equation}
 a=\max\{j\in\{0,\ldots,k\}:\tilde t_\ell=g_\ell
       \text{ for all }1\leq\ell\leq j\}.
 \label{eq:verified-prefix}
\end{equation}
The decoder accepts $\tilde t_{1:a}$ and
emits $g_{a+1}$. At a mismatch, later candidates are discarded; subsequent
proposals use the updated generated prefix.}

We reuse text rather than KV, because adding evidence changes the conditioning context.
Implementation details appear in
Appendix~\ref{app:correction-implementation}.

\subsection{Gate lookahead with answer-version validation}
\label{sec:gate-lookahead}

\noindent{Text reuse reduces decoding iterations within a call.
We further reduce waiting between correction steps by allowing the next
check to run before the preceding check and any resulting revision finish.
In Figure~\ref{fig:reuse}c, the year check and location check start on the
same answer $A_0$. If the year step leaves the answer unchanged, the early
location result can be used when its turn arrives, avoiding a separate
wait for that check. In the illustrated case, however, the editor changes
1997 to 1998 and produces $A_1$. The location check must then run again on
$A_1$: its complete input has changed even though the edited fact concerns
the year.}

{After retrieval, our implementation keeps up to two checks in
flight while applying decisions and revisions in the original
query/evidence order. Each early check records the answer and complete
prompt it used. Before adopting its result, \workname{} requires both to
match the input that the sequential workflow would use at that position.
A matching result is
reused; otherwise, the check is repeated on the updated answer. An edit
that leaves the answer unchanged, or is rejected by the existing revision
policy, does not invalidate the next check.}

{These two mechanisms operate at different levels: verified text
reuse shortens each gate or editor call, while gate lookahead overlaps
checks across calls. 
Further implementation details are provided in
Appendix~\ref{app:correction-implementation}.}

\subsection{Partial decoupling of attention execution}
\label{sec:partial}

\noindent A shared model forward can contain prompt processing
for new questions or correction calls, single-token decoding, and multi-token
candidate verification. Computation sharing and request completion also
change the number of requests in each group. An attention execution path
suited to a large single-token group may therefore be inefficient when
longer queries are mixed in or the group becomes small. We address these
differences by grouping attention queries and selecting how each group
executes, while retaining joint batching for linear layers.

For one forward pass, let $\mathcal P$ denote prefill requests,
$\mathcal D$ non-prefill single-token decoding requests, and $\mathcal C$ multi-token
verification requests. A prefill request $p$ processes $q_p$ prompt tokens
in this pass, including when the prompt is processed in chunks; a decoding
request processes one token, and a verification request $c$ processes
$q_c>1$ tokens. Requests sharing the same forward computation are counted
only once. Together, they contribute
\begin{equation}
 M=\sum_{p\in\mathcal P}q_p+|\mathcal D|+\sum_{c\in\mathcal C}q_c
 \label{eq:rows}
\end{equation}
token rows to the batched Q/K/V projection (Figure~\ref{fig:execution}).
Only attention is partitioned: sub-batches execute sequentially,
each using its requests' own KV caches and causal visibility. Their outputs
return to the original row order before the batched output projection and
FFN. Grouping metadata is prepared once per step and reused across layers.

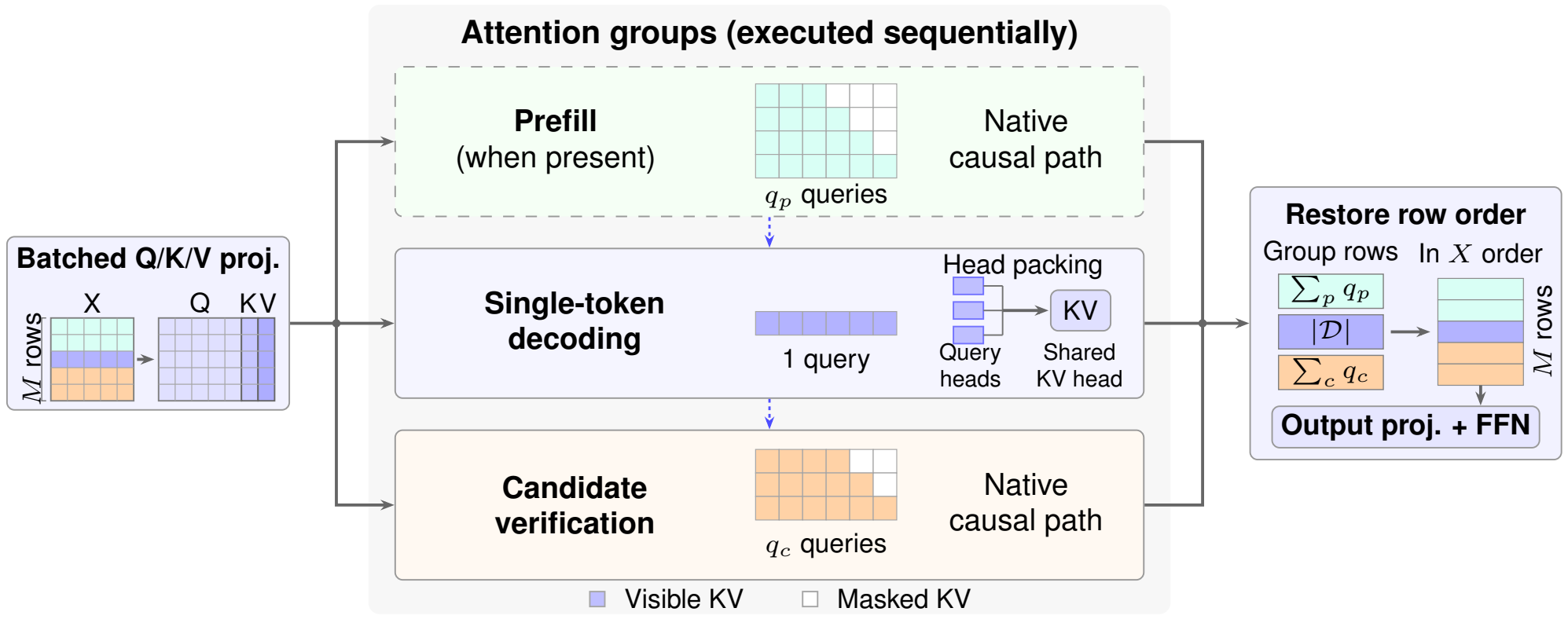
\begin{figure}[t]
\centering
\scalebox{0.9}{%
\begin{tikzpicture}[x=1cm,y=1cm,font=\sffamily\fontsize{8}{9.5}\selectfont,
 box/.style={draw=black!40,rounded corners=2pt,align=center,inner sep=3pt},
 sidetitle/.style={font=\sffamily\bfseries\fontsize{7.5}{8.5}\selectfont,
   anchor=base,inner sep=0pt,yshift=-.35cm},
 arr/.style={-{Stealth[length=1.3mm]},thick,draw=black!60},
 orderarr/.style={-{Stealth[length=1.2mm]},draw=blue!70,line width=.6pt,
   dash pattern=on 1pt off 1pt},
 tinylabel/.style={font=\sffamily\fontsize{7}{8}\selectfont,align=center}]
\useasboundingbox (-.6,0) rectangle (14.3,5.85);

\fill[black!3,rounded corners=3pt] (3.1,.08) rectangle (10.6,5.78);
\node[font=\sffamily\fontsize{8.5}{10}\selectfont\bfseries]
 at (6.85,5.5){Attention groups (executed sequentially)};

\def\inputrowbands{0/green!25!cyan!15,1/green!25!cyan!15,2/blue!30,3/orange!35,4/orange!35}
\node[box,fill=blue!5,minimum width=2.64cm,minimum height=1.62cm]
 (qkv) at (1.04,2.8) {};
\node[sidetitle,yshift=.05cm] at (qkv.north){Batched Q/K/V proj.};
  \begin{scope}[shift={(.13,2.23)},x=.155cm,y=.155cm,
    every node/.style={tinylabel,inner sep=0pt}]
    \node at (2.5,4.9) {X};
    \node at (9,4.9) {Q};
    \node at (11.9,4.9) {K};
    \node at (13.1,4.9) {V};
    \node[rotate=90] at (-1.15,1.5) {$M$ rows};
    \draw[draw=black!55,line width=.3pt]
      (-.28,-1) -- (-.55,-1) -- (-.55,4) -- (-.28,4);
    \foreach \r/\shade in \inputrowbands
      \path[fill=\shade] (0,{4-\r}) rectangle (5,{3-\r});
    \draw[draw=black!50] (0,-1) rectangle (5,4);
    \draw[arr] (5.2,1.5) -- (6.25,1.5);
    \path[fill=blue!12,draw=black!50] (6.5,-1) rectangle (11.5,4);
    \path[fill=blue!22,draw=black!60] (11.5,-1) rectangle (12.5,4);
    \path[fill=blue!32,draw=black!60] (12.5,-1) rectangle (13.5,4);
    \foreach \x in {1,...,4}{
      \draw[draw=black!35,line width=.25pt] (\x,-1) -- (\x,4);
      \draw[draw=black!35,line width=.25pt] ({6.5+\x},-1) -- ({6.5+\x},4);
    }
    \foreach \y in {0,...,3}{
      \draw[draw=black!35,line width=.25pt] (0,\y) -- (5,\y);
      \draw[draw=black!35,line width=.25pt] (6.5,\y) -- (13.5,\y);
    }
  \end{scope}
\node[box,fill=blue!5,text width=2.7cm,minimum height=2.55cm]
 (out) at (qkv.center -| 12.8,0) {};
\node[sidetitle] at (out.north){Restore row order};
\begin{scope}[shift={(out.center)},groupcount/.style={draw=black!50,line width=.3pt,
 minimum width=.98cm,minimum height=.32cm,inner sep=0pt,tinylabel}]
 \node[tinylabel,inner sep=0pt] at (-.70,.66){Group rows};
 \node[tinylabel,inner sep=0pt] at (.70,.66){In $X$ order};
 \node[groupcount,fill=green!25!cyan!15] at (-.70,.30){$\sum_p q_p$};
 \node[groupcount,fill=blue!30] at (-.70,-.08){$|\mathcal D|$};
 \node[groupcount,fill=orange!35] at (-.70,-.46){$\sum_c q_c$};
 \draw[arr] (-.14,-.08) -- (.23,-.08);
 \foreach \r/\shade in \inputrowbands
   \path[fill=\shade,draw=black!45,line width=.25pt]
     (.30,{.42-.20*\r}) rectangle (1.10,{.42-.20*(\r+1)});
 \node[tinylabel,rotate=90,inner sep=0pt] at (1.28,-.08){$M$ rows};
 \node[box,fill=blue!10,minimum width=2.50cm,inner sep=2pt,
   font=\sffamily\bfseries\fontsize{7.5}{8.5}\selectfont]
   (ffn) at (0,-.97){Output proj. + FFN};
 \coordinate (lastrow) at (.70,-.58);
 \draw[arr] (lastrow) -- (ffn.north -| lastrow);
\end{scope}

\node[box,dashed,fill=green!4,minimum width=7cm,minimum height=1.4cm]
 (prefill) at (6.85,4.5) {};
\node[box,fill=blue!4,minimum width=7cm,minimum height=1.4cm]
 (single) at (6.85,2.8) {};
\node[box,fill=orange!5,minimum width=7cm,minimum height=1.4cm]
 (multi) at (6.85,1.1) {};

\draw[orderarr] (prefill.south) -- (single.north);
\draw[orderarr] (single.south) -- (multi.north);

\node[align=center] at (prefill.center -| 4.85,0)
 {\textbf{Prefill}\\(when present)};
\node[align=center] at (single.center -| 5.03,0)
 {\textbf{Single-token}\\\textbf{decoding}};
\node[align=center] at (multi.center -| 5.03,0)
 {\textbf{Candidate}\\\textbf{verification}};

\begin{scope}[shift={(6.72,5.04)},x=.22cm,y=-.22cm]
 \foreach \r in {0,...,3}{
  \foreach \c in {0,...,5}{
   \pgfmathtruncatemacro{\visible}{\c <= \r+2}
   \ifnum\visible=1
    \path[fill=green!25!cyan!15,draw=black!35,line width=.25pt]
     (\c,\r) rectangle ({\c+1},{\r+1});
   \else
    \path[fill=white,draw=black!35,line width=.25pt]
     (\c,\r) rectangle ({\c+1},{\r+1});
   \fi
  }
 }
\end{scope}
\node[tinylabel] at (7.38,3.98){$q_p$ queries};
\begin{scope}[shift={(6.72,2.91)},x=.22cm,y=-.22cm]
 \foreach \c in {0,...,5}{
  \path[fill=blue!30,draw=black!35,line width=.25pt]
   (\c,0) rectangle ({\c+1},1);
 }
\end{scope}
\node[tinylabel] at (7.38,2.45){1 query};
\begin{scope}[shift={(6.72,1.62)},x=.22cm,y=-.22cm]
 \foreach \r in {0,...,2}{
  \foreach \c in {0,...,5}{
   \pgfmathtruncatemacro{\visible}{\c <= \r+3}
   \ifnum\visible=1
    \path[fill=orange!35,draw=black!35,line width=.25pt]
     (\c,\r) rectangle ({\c+1},{\r+1});
   \else
    \path[fill=white,draw=black!35,line width=.25pt]
     (\c,\r) rectangle ({\c+1},{\r+1});
   \fi
  }
 }
\end{scope}
\node[tinylabel] at (7.38,.72){$q_c$ queries};
\node[align=center] at (prefill.center -| 9.25,0){Native\\causal path};
\node[align=center] at (9.25,1.1){Native\\causal path};

\node[tinylabel] at (9.22,3.33){Head packing};
\foreach \y in {2.69,2.92,3.15}{
 \draw[draw=blue!55,fill=blue!25] (8.57,\y-.08) rectangle (8.85,\y+.08);
 \draw[draw=black!55] (8.85,\y) -- (9.03,\y);
}
\draw[draw=black!55] (9.03,2.69) -- (9.03,3.15);
\draw[arr] (9.03,2.92) -- (9.46,2.92);
\node[box,fill=blue!12,minimum width=.56cm,minimum height=.38cm,
 inner sep=1pt,tinylabel] at (9.76,2.92){KV};
\begin{scope}[every node/.append style={align=center,font=\sffamily\fontsize{6}{7}\selectfont}]
 \node at (8.73,2.41){Query\\heads};
 \node at (9.75,2.41){Shared\\KV head};
\end{scope}

\draw[thick,draw=black!60] (qkv.east) -- (2.8,2.8);
\draw[arr] (2.8,2.8) |- (prefill.west);
\draw[arr] (2.8,2.8) -- (single.west);
\draw[arr] (2.8,2.8) |- (multi.west);
\fill[black!60] (2.8,2.8) circle (1.1pt);
\draw[thick,draw=black!60] (prefill.east) -- (10.9,4.5) -- (10.9,2.8);
\draw[thick,draw=black!60] (single.east) -- (10.9,2.8);
\draw[thick,draw=black!60] (multi.east) -- (10.9,1.1) -- (10.9,2.8);
\fill[black!60] (10.9,2.8) circle (1.1pt);
\draw[arr] (10.9,2.8) -- (out.west);

\draw[draw=black!35,fill=blue!25] (5.17,.15) rectangle +(.14,.14);
\node[tinylabel,anchor=west] at (5.37,.22){Visible KV};
\draw[draw=black!35,fill=white] (7.16,.15) rectangle +(.14,.14);
\node[tinylabel,anchor=west] at (7.36,.22){Masked KV};
\end{tikzpicture}
}
\caption{Partial decoupling within one model forward.
The Q/K/V projection processes all $M$ token rows together.
In mixed batches, only attention is split into sequential calls, one for
each group present. The attention outputs return to their original row
positions before the output projection and FFN process all rows together.}
\label{fig:execution}
\end{figure}

FlashAttention-2's single-token decode path~\citep{dao2024flashattention2}
requires a maximum query length of one across the call. A single
multi-token prefill or verification request raises that maximum for the
whole batch. As Figure~\ref{fig:execution} shows, \workname{} runs a separate
attention call for each group present in a mixed batch.
\emph{Prefill} processes the prompt tokens scheduled for this step.
\mbox{\emph{Single-token decoding}} batches one query token from each decoding
request, and \mbox{\emph{Candidate verification}} processes the query positions in
each multi-token candidate block. 
With the longer queries in separate calls, the decode
call has a maximum query length of one.

Under grouped-query attention (GQA)~\citep{ainslie2023gqa}, several
query heads within each decode request attend to the same cached keys and
values. For small decode groups without prefill, \workname{} passes
the queries directly to FlashAttention-2, which packs these heads internally.
For larger groups, \workname{} rearranges the queries before the
attention call so that heads sharing a KV head are processed together.
After the call, \workname{} restores the outputs to their original head order.
Appendix~\ref{app:attention-implementation} gives the eligibility rules
and the kernel settings used by each path.

\section{Experiments}
\label{sec:experiments}

{In the experimental study, we evaluate JCT,
throughput, and GPU execution time. We also assess whether detector decisions
and final answers are preserved.} The main performance study uses 256 questions with
32 continuously replenished workflows. The cumulative ablation
uses the same 256-question workload.
{All experiments are conducted on a single NVIDIA RTX A6000 GPU.}
Appendix~\ref{app:measurement} details the measurement windows
and numerical controls.

\subsection{Workload, controls, and measurement}

\paragraph{Complete workflow.}
We use HotpotQA questions \citep{yang2018hotpotqa} and a frozen corpus of
269,612 passages, with {BM25 retrieval~\citep{robertson2009bm25}
and a MiniLM reranker~\citep{wang2020minilm}}. RARR prompts
and ordered checks/edits are adapted to local retrieval and a shared
Qwen2.5-14B-Instruct model~\citep{qwen2025qwen25}. Short answers use a one-complete-sentence
instruction and 256-token target/auxiliary caps; long answers use a 120-word paragraph instruction and 512-token caps. 
Each question has ten independent auxiliaries. Three independent qgen calls
use temperature 0.7 and a 512-token cap; gates and editors use greedy
structured generation. {CoCoA routing thresholds are
$0.083$ (short) and $0.097$ (long), calibrated on NQ and SQuAD, respectively,
at a 10\% false-positive rate using a lexical correctness proxy.
We transfer them unchanged to all HotpotQA configurations.}

The engine uses BF16, vLLM 0.31.0 V2 asynchronous scheduling, FA2, APC,
chunked prefill, and supported CUDA Graphs, with 384 sequence slots,
a 2,048-token budget, context 8,192, and GPU memory utilization 0.8.
Workflow concurrency is 32, with eight CPU retrieval workers.

\paragraph{Configurations.}
\textbf{{Native vLLM (Native)}} {runs the reference
detection--correction workflow. It generates all ten auxiliary responses,
runs the three query-generation calls concurrently, and performs correction
checks sequentially.}
\textbf{{\workname{}}} adds 
detection and query sharing, certified early stopping, correction speculation, attention decoupling, and gate lookahead.

\paragraph{{Numerical settings.}}
Our main performance comparison uses
vLLM's \textbf{default} numerical settings.
Under continuous batching,
the same request can be grouped with different requests across runs. Even
with a fixed sampling seed, changes in batch composition can affect
floating-point results and token selection. To control for this variation,
we run both systems with vLLM's \textbf{batch-invariant} setting\footnote{\url{https://docs.vllm.ai/en/v0.31.0/features/batch_invariance/}}
in a separate experiment and check whether \workname{} reproduces Native's
detector decisions and final answers for each question.

\paragraph{{Workloads and evaluation sets.}}
Our main evaluation uses 256 HotpotQA questions with up to 32
concurrent detection--correction workflows. Each completed workflow is
replaced by a new question until all 256 questions have been submitted.
The cumulative ablation uses the same questions and replenishment policy.
The held-out evaluation uses a disjoint set of 256 HotpotQA questions with the
same concurrency and replenishment policy. 

\paragraph{Measurement.}

Throughput and JCT are measured over a central completion window, and GPU time over the full workload (Appendix~\ref{app:load-accounting}). For the main and held-out comparisons, we report the mean of four timing runs and three Nsight Systems profiles per configuration and answer length; the cumulative ablation uses three runs and two profiles. p95/p99 JCT are averaged over per-run quantiles, and GPU time is the union of GPU kernel intervals. Timings exclude model loading and warmup.

\subsection{Complete pipeline under replenished load}

\begin{table}[t]
\centering
\caption{Native versus \workname{}:
256 questions, 32 continuously replenished workflows. 
GPU time averages three independent
whole-workload profiles.}
\label{tab:load-default}
\vspace{3pt}
\includegraphics[width=0.9\linewidth]{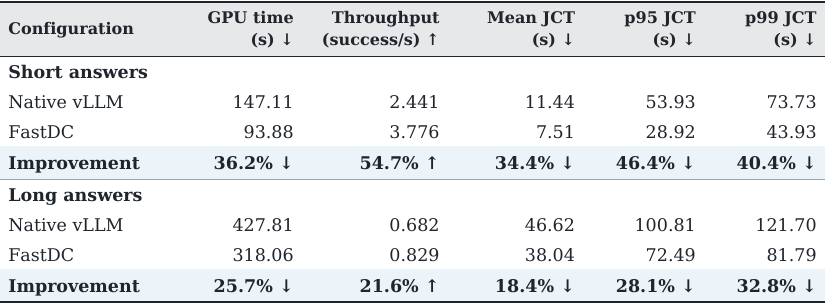}
\end{table}

As shown in Table~\ref{tab:load-default}, \workname{} improves throughput and JCT under continuously
replenished load, with larger gains for short answers. 
Throughput reaches $1.55\times$ Native for short answers and
$1.22\times$ for long answers, while whole-workload GPU time falls by
36.2\% and 25.7\%, respectively. Short-answer mean, p95, and p99 JCT
fall by 34.4\%, 46.4\%, and 40.4\%, respectively; the corresponding
long-answer reductions are 18.4\%, 28.1\%, and 32.8\%.
Throughput and mean JCT improve in all four reported timing repeats.

\workname{} achieves a larger speedup on short answers in terms of GPU time and throughput.
This is because more computation can be shared across samples in the short answer scenario. According to the GPU profiles,
prefix sharing avoids duplicate computation for 29.3\% of logical token rows
in the short-answer workload, compared with 6.7\% in the long-answer workload.
When the detector flags an answer, the workflow must go to the correction stage and check the generated result
before returning the final answer. In the timing runs, only
14.4\% of short-answer requests required this correction step, compared
with 55.5\% of long-answer requests. This makes the shared sampling advantage more significant in the short answering scenario.

\subsection{{Cumulative Ablation}}

We now proceed to the ablation study. Figure~\ref{fig:cumulative-ladder} evaluates seven cumulative
configurations, using 256 questions and 32 continuously
replenished workflows. 
We average throughput over three runs per configuration and answer length. 
Figure~\ref{fig:cumulative-ladder} adds one mechanism per row, on the same 256-question workload.

\begin{figure}[t]
\centering
\includegraphics[width=\linewidth]{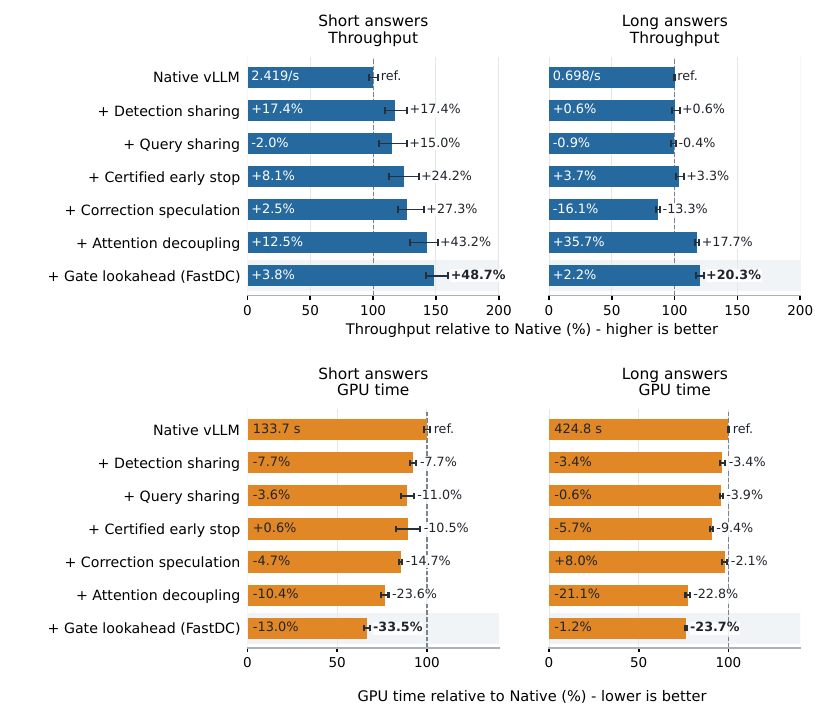}
\caption{Cumulative ablation: 256 questions,
32 continuously replenished workflows.
Bars are normalized to Native; inner labels show the increment over the row above, outer labels the cumulative change. Whiskers: min--max over runs.
}
\label{fig:cumulative-ladder}
\end{figure}

\paragraph{{Shared sampling and early stopping.}}
Detection sharing improves short-answer throughput by 17.4\%
and reduces GPU time in both answer conditions; its long-answer throughput
benefit is smaller. As mentioned before, this happens since long answering scenario results in lower shared sampling rate. Query-generation sharing adds modest average GPU
reductions but no further throughput improvement. Early
stopping then increases short-answer throughput by 8.1\%, although its
GPU increment has opposite signs in the two profiles. 
For short answers, early stopping raises throughput by $8.1\%$ without a consistent GPU saving, since it mainly shortens detection latency; for long answers, it also reduces GPU time by $5.7\%$.

\paragraph{Correction speculation, attention decoupling, and gate lookahead.}
Correction speculation slightly improves short-answer throughput,
but reduces long-answer throughput by 16.1\% and increases GPU time by
8.0\%. This is because long answers enter correction more often, so more requests undergo
speculative verification. When multi-token verification and single-token
decoding share an attention call, the decoding requests lose access to the
optimized single-token path. In the long-answer profiles, speculation
reduces the number of model forward passes, yet total attention kernel
time increases. With attention decoupling, long-answer throughput rises by
35.7\% and GPU time falls by 21.1\% relative to the preceding configuration
with mixed attention.
Gate lookahead increases throughput by 3.8\% for short answers and
2.2\% for long answers, while reducing short-answer GPU time by 13.0\%.

Overall, sampling sharing yields the largest incremental throughput gain for short answers, while attention decoupling yields the largest gain for long answers.

\subsection{Performance on held-out questions}

\begin{table}[t]
\centering
\caption{Performance on 256 held-out HotpotQA questions. 
We report throughput (successful answers/s) and JCT over
four runs per condition, and whole-workload GPU time over three independent
profiles.}
\label{tab:heldout}
\vspace{3pt}
\includegraphics[width=0.9\linewidth]{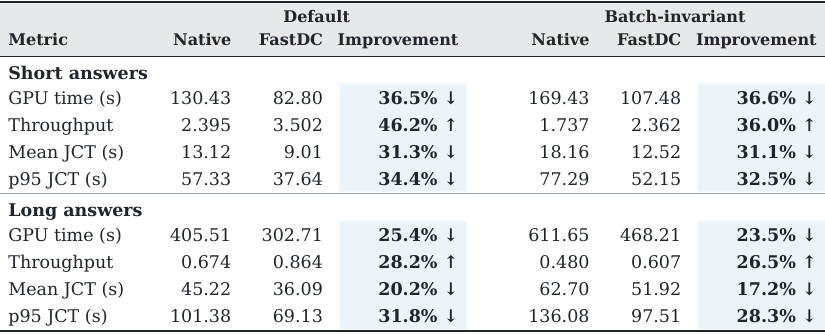}
\end{table}

To test whether the performance gains extend to other questions,
we repeat the main comparison on 256 additional HotpotQA questions that do
not overlap with those in Table~\ref{tab:load-default}, using the same
workload settings. With default execution, \workname{} improves throughput
over Native by 46.2\% for short answers and 28.2\% for long answers
(Table~\ref{tab:heldout}). On both question sets,
\workname{} reduces GPU time by about 36\% for short answers and
25--26\% for long answers (Tables~\ref{tab:load-default} and~\ref{tab:heldout}).
These similar savings show that \workname{}'s GPU efficiency gains carry
over to a new set of HotpotQA questions.

The gains also persist when both systems use batch-invariant
execution on the same held-out questions. \workname{} increases throughput
by 36.0\% for short answers and 26.5\% for long answers, with higher
throughput than Native in all four runs for each answer length.
Mean JCT falls by 31.1\% and 17.2\%, respectively.

\subsection{Do detector decisions and final answers match?}
\label{sec:preservation-results}

Under default execution, 
\workname{} and Native made the same
decision about whether to start correction on 96.5\% of short-answer requests
and 93.0\% of long-answer requests (Table~\ref{tab:load-outcomes}).
Among requests for which both systems returned an answer, the final texts
matched exactly in 80.0\% of short-answer cases and 24.0\% of long-answer
cases. The 24.0\% match rate falls within the range we observed when
Native repeated the same workload.
Comparing the
first Native run with each of the next three, we found that only
22.5--26.9\% of long answers returned in both runs matched exactly. The lower match percentage of long answers compared with short answers suggests the higher uncertainty of longer language generation.

\begin{table}[htbp]
\centering
\caption{Output matches on the held-out questions in
Table~\ref{tab:heldout}. Each row pools four runs on the same 256 questions.
We compare the decision to start correction for all 1,024 request pairs.
We compare final texts for exact matches when both systems return an answer.}
\label{tab:load-outcomes}
\vspace{3pt}
\includegraphics[width=0.9\linewidth]{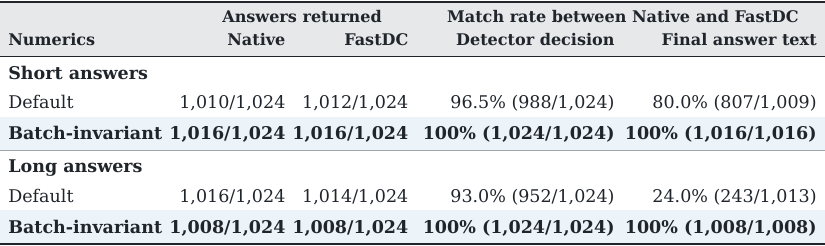}
\end{table}

To test whether \workname{} preserves Native's outputs while
controlling for numerical variation across batches, we repeat the comparison
with batch-invariant execution. \workname{} and Native produced identical
workflow outputs in all four runs. Their detector decisions matched on all
256 questions for both answer lengths. In each run, both systems returned
answers to the same 254 questions in the short-answer setting and the same
252 in the long-answer setting. For every question they answered, the final
texts matched exactly.
This batch-invariant evaluation shows that \workname{} preserves Native's
detector decisions and final answers on the held-out questions.

\Needspace{5\baselineskip}
\subsection{{Mechanism Analysis}}

\paragraph{Shared sampling and text reuse.}
As shown in Table~\ref{tab:mechanism-analysis}(a), shared sampling merges 29.3\% of logical token
rows for short answers and 6.7\% for long answers.
We replay the n-gram lookup on
the check and edit calls recorded in these same 256-question runs.
Candidates drawn from the answer under revision match 38.3\% of the
editor's output tokens for long answers and 26.8\% for short answers.
Most of these matches appear in the revised answer itself. Correcting
one fact in a paragraph can leave whole sentences intact, giving the
editor more text to reuse than in a single-sentence answer.

\begin{table}[!htbp]
\centering
\caption{Reuse by answer length (a) and per-layer attention time by request grouping (b).}
\label{tab:mechanism-analysis}
\vspace{3pt}
\includegraphics[width=0.9\linewidth]{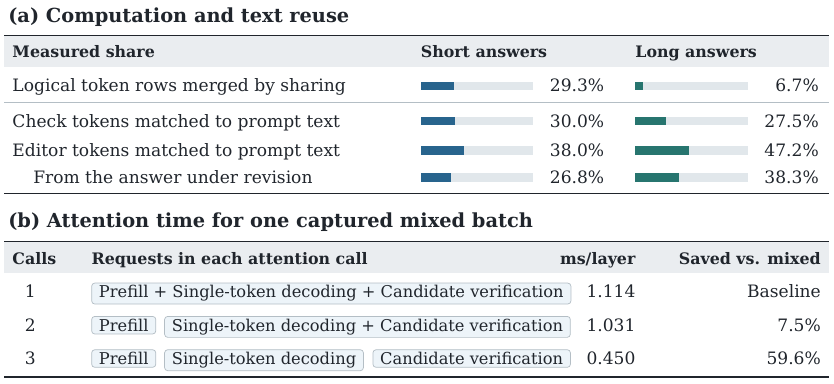}
\end{table}

\paragraph{Attention decoupling.}
We replay the same 5 prefill, 118 single-token decoding, and 2 candidate verification
requests in all three layouts, holding Q/K/V fixed at each of layers 0 and
23 (Table~\ref{tab:mechanism-analysis}(b)). Separating
prefill alone saves 7.5\%. Giving decoding its own call also enables head
packing, reducing attention time from 1.114 to 0.450\,ms per layer,
including sequential calls and row reordering. In this batch, separating
single-token decoding from candidate verification accounts for about
88\% of the total attention time saved.

\section{Related Work}
\label{sec:related}

\paragraph{Detection and evidence-based revision.}
When an LLM is asked for a year, sampled responses such as
``1998'' and ``It was in 1998'' express the same answer.
\citet{kuhn2023semantic} group such paraphrases and compute entropy over
the probabilities of the distinct meanings. This semantic entropy rises
when probability spreads across different answers, such as 1997 and 1998.
\citet{farquhar2024detecting} use high semantic entropy to flag possible
hallucinations.
SelfCheckGPT checks whether other passages sampled for the same prompt support
each sentence in the original response~\citep{manakul2023selfcheck}.
CoCoA combines the model's confidence in the original answer, estimated from
token probabilities, with how closely that answer matches the other samples in
meaning~\citep{vashurin2025cocoa}. Our end-to-end evaluation uses the CoCoA
score studied by \citet{pawar2026tokens} to decide which answers need correction.
For correction, RARR retrieves evidence for claims in the answer and revises
unsupported claims while preserving as much of the answer as
possible~\citep{gao2023rarr}. \workname{} serves the detector and the RARR
revision steps on a shared model, using a fixed local corpus for retrieval.

\paragraph{Sampling and computation sharing.}
Adaptive semantic-entropy estimation \citep{sun2026efficient} and
Adaptive-Consistency \citep{aggarwal2023adaptive} allocate samples for
estimation or answer selection. \workname{} can stop sampling early
and still return the decision the detector would reach with all $N$ samples.
PagedAttention
manages paged KV sharing \citep{kwon2023pagedattention},
SGLang reuses cached prefixes \citep{zheng2024sglang}, and Hydragen separates
shared-prefix from suffix attention \citep{juravsky2024hydragen}.
There is a concurrent study in reasoning search. SpecScale computes the next-token distribution
once for candidates with the same token prefix and samples independently for
each candidate~\citep{jeong2026specscale}. \workname{} applies the same
principle to the auxiliary responses used for hallucination detection and
the query plans used for correction.

\paragraph{Speculative decoding for correction.}
Speculative decoding verifies inexpensive proposals with a target model
\citep{leviathan2023speculative}. During correction, \workname{}
uses n-gram lookup to propose continuations from the current prompt or
from text already generated in the same call. The model checks the proposed
tokens in one forward pass and accepts them up to the first mismatch.

\paragraph{Serving interference.}
SmartSpec selects speculation length using goodput and observed acceptance
\citep{liu2024smartspec}, including high-load regimes where speculation can
become counterproductive. Sarathi-Serve balances
prefill and decode through chunking and token budgets
\citep{agrawal2024sarathi}. FlashAttention optimizes attention execution
\citep{dao2022flashattention}. These works establish the broader problems of
heterogeneous query cost, tile effects, and serving-aware speculation.
\workname{} complements engine scheduling with operator-level partitioning:
single-query detection and multi-query correction use separate attention
paths while retaining joint matrix multiplications. 

\section{{Conclusion}}
\label{sec:discussion}

In this paper, we present \workname{}, a unified system for
efficient sampling-based hallucination detection and correction in vLLM.
It shares model computation across independent samples, reuses existing
text through verified speculation, and overlaps correction checks while
preserving revision order. It partitions attention for the combined
workload while retaining jointly batched linear layers.
Experiments with Qwen2.5-14B on an RTX A6000 show that
\workname{} achieves $1.55\times$ and $1.22\times$ the throughput of native
vLLM for short and long answers, respectively, while reducing GPU execution
time by 36.2\% and 25.7\%.
{Our current design targets single-GPU execution. Future work
will explore distributed serving, including how to place and schedule
detection and correction across GPUs while balancing computation reuse,
load distribution, and communication costs.}

\bibliography{iclr2026_conference}
\bibliographystyle{iclr2026_conference}
\clearpage
\appendix
\section{{Shared-Sampling Implementation Details}}
\label{app:shared-implementation}

{\paragraph{Native request and KV management.}
vLLM retains responsibility for KV allocation, prefix caching, preemption,
and reclamation. Within an explicit sampling group, the adapter selects one
representative for requests with identical complete token histories,
computed positions, and current scheduled spans. It copies newly computed
KV entries into the other requests' native allocations and expands the
representative outputs to one row per logical request before native grammar,
sampling, and log-probability processing. Each sample retains its identity,
seed, and sampling position independently of representative selection.
Repeated responses remain separate logical samples. KV allocations remain
separate rather than forming a compressed prefix tree.}

{\paragraph{Prefill eligibility.}
Automatic prefix caching reuses available full prompt blocks. The guarded
prefill path additionally shares eligible final spans inside a partial
prompt block and their first-token computation. Full-block writers remain
active because other scheduled requests may read those blocks during the
same forward. These checks limit which prefill work can be shared; they do
not remove all duplicate prefill computation.}

{\paragraph{Asynchronous state validation.}
The V2 engine schedules asynchronously, so its scheduled-progress counters
can be ahead of the state already computed on the GPU. The adapter reads
actual computed positions and histories before grouping requests, and also
checks that their current scheduled spans match. The implementation includes
synchronization and CPU history comparison; both costs are included in
service measurements.}

{\paragraph{Query-generation output assembly.}
All three query-generation calls run concurrently. Their outputs are
collected by fixed logical index before the unchanged query union is formed.
The native baseline uses the same three-way concurrency, so this concurrency
is not attributed to shared sampling.}

\section{CoCoA Certificate Details}
\label{app:bounds}

In Eq.~\eqref{eq:cocoa-bound}, $A$ is the mean negative
log-probability of the tokens in the original answer. Clamping cosine
similarity to $[-1,1]$ bounds each auxiliary response's distance from that
answer between 0 and 2. Once the original answer is complete, the score
therefore lies in $[0,2A]$ even before any auxiliary response is scored.
Auxiliary responses that finish first are buffered until then.

\workname{} uses a margin of $\epsilon=10^{-10}$ near the
threshold $\tau$. It returns a low-risk decision if $U_S+\epsilon\leq\tau$
and a high-risk decision if $L_S>\tau+\epsilon$. While the decision is
uncertain, each newly scored auxiliary response updates the bounds.
If all $N$ responses are needed, the detector applies its original
threshold to the full score.

{A generation producer continuously reads vLLM's output streams
and emits a completion event for each logical sample. A separate consumer
performs CPU semantic encoding, reuses embeddings for identical texts, and
updates the score bounds. Generation continues during encoding, and the
consumer processes completed responses without waiting for a full sampling
wave. Once the decision is fixed, the workflow submits native abort requests
for unfinished auxiliaries and waits for cleanup. Failure handling releases
all owned requests and admission slots.}

{Cancellation preserves the original fixed-budget threshold
decision. It prevents further generation steps from being scheduled after
the engine processes the abort request; it cannot recover computation already
performed or stop an in-flight GPU step.}
Figure~\ref{fig:cumulative-ladder} measures the increment from adding
certified cancellation after detection and qgen sharing. The certificate
guarantees the threshold decision; its performance effect depends on the
remaining work and the cost of processing completions and cancellations.

\section{{Correction Execution Details}}
\label{app:correction-implementation}

{\paragraph{Candidate lookup and verification.}
Native n-gram prompt lookup supports the greedy structured gate and editor
calls. Matching suffixes of length four to eight can propose up to eight
subsequent tokens from the current prompt or generated history. The LLM
verifies candidates under the call's current prompt and native output-format
constraints. Candidate text may contain factual errors; acceptance requires
agreement with the model's greedy predictions, rather than a claim that
the candidate source is factually correct. The n-gram algorithm is an
existing engine primitive.}

{\paragraph{Verification widths and graph execution.}
If the scheduler reserves $r$ candidate positions but the proposer provides
$v$, the worker evaluates $\min(r,v)+1$ rows, including the normal token.
Native logical reservation, grammar mapping, and rollback remain consistent
with the executed span. This removes empty verification without assuming
future acceptance. A graph selection fix retains compatible single-token
captures alongside speculative captures; it does not provide FULL graph
replay for arbitrary mixed widths. Detection/query-generation sharing and
correction speculation coexist in one engine. Speculative requests are not
merged using the ordinary shared-state rule.}

{\paragraph{Lookahead lifecycle and preservation.}
The logical seed, output constraints, evidence, and generation settings of
each check are fixed by its position in the correction sequence. If an
accepted revision changes the answer, completed checks from the old answer
are discarded and unfinished attempts are canceled through native abort.
The workflow waits for cleanup and reissues those checks on the updated
answer using the same logical seeds. No stale verdict or KV is transplanted
into the new prompt. Total device time includes useful and discarded
attempts. Preservation checks enforce the sequential workflow's inputs and
revision order; experiments with batch-invariant numerics additionally check outputs
under the changed batching order.}

\section{{Attention Decoupling Implementation Details}}
\label{app:attention-implementation}

\paragraph{Sub-batch construction and KV mapping.}
The backend prepares group metadata once per step after shared sampling
and reuses it across layers. For mixed-prefill batches, native prefill flags
identify prompt work; actual query lengths distinguish the remaining
single-token decoding and multi-token verification requests. Grouping does
not depend on whether a request belongs to detection or correction. Each group retains
the corresponding requests' context lengths and KV block-table entries;
requests in one group need not have identical prompts or context lengths.
KV updates retain one native write and the original slot mapping.
Contiguous groups use tensor views; noncontiguous groups use gather/scatter,
whose cost is included. The prefill and small-group paths derive each group's
maximum context-length bound from native CPU metadata without an additional
device synchronization. The attention groups execute sequentially, and
their outputs return to the original row and head order before the next
mixed linear layer. The groups have no separate request queues.

\paragraph{GQA packing and causal visibility.}
In the evaluated Qwen model, five query heads share each of eight KV heads.
Packing rearranges Q from $[n,40,128]$ to $[5n,8,128]$ within the $n$ requests.
The packed heads of each request represent the same final position, so the
decode call uses noncausal attention over that request's permitted KV prefix.
Prefill and verification retain their causal masks. For eligible small groups,
FlashAttention-2 performs the same head rearrangement internally.

\paragraph{Numerical settings and validation.}
In our default-mode probes, the mixed call used \texttt{num\_splits=0} and
128-position KV tiles. Given the same setting, FlashAttention-2's automatic
packing path selected one split and 64-position tiles. Packing Q before the
call preserved the mixed call's split setting and tile size. Tile size affects
the reduction order and can change BF16 outputs. Both paths pass one split
under batch-invariant numerics. In a separate default-mode replay, we
compared the mixed call with explicit head packing after partitioning.
Their attention outputs matched element by element on identical Q/K/V inputs
from three captured batches at each of layers 0, 23, and 47.

\paragraph{Attention path selection.}
\workname{} selects the attention path using fixed rules based on the
presence of prefill, the sizes of the decoding and verification groups,
and their query and context lengths.
The packed decode/verification path requires at least 32 single-token
requests after sharing, at least one multi-token verification request,
and maximum query length at most nine. When larger prefill spans are present,
the mixed-prefill extension separates prefill, single-token decoding, and
verification, provided at least 32 non-prefill single-token requests remain.
For a batch without prefill, let $d$ be the number of single-token
requests after sharing. The small-group path requires $8\le d\le31$ and
between 1 and $\lfloor d/2\rfloor$ verification requests, each processing at
most nine tokens. In each group, the largest context-length upper bound
in the CPU metadata must lie between 1,024 and 4,096 tokens. In replays of small batches with
identical Q/K/V inputs, the two native calls and row reordering together
took less time than the mixed call. Batches that meet none of the splitting
rules use the native attention call.

\section{Measurement Windows and Numerical Controls}
\label{app:measurement}

{Native and \workname{} use the same instrumented vLLM installation
and both reuse CPU embeddings for identical response texts.}

\subsection{Workload and window accounting}
\label{app:load-accounting}

The main and held-out evaluations each use 256 HotpotQA questions
selected using only source IDs and question text. The sets are disjoint
by source ID and normalized question text. The
cumulative ablation uses the main set. In all three studies, each completed
or failed workflow releases a slot for the next question in the fixed
order until all 256 questions have been submitted.

{Whole-workload GPU time includes pipeline startup and drain.
Throughput is the number of successful completions between completion
counts 64 and 192 divided by the elapsed time; completion counts include
both successes and failures. JCT uses zero-based arrival indices
64--191 and tracks each selected request through completion, including
requests that finish after the throughput window ends.}

\subsection{Numerical controls}
\label{app:numerical-controls}

In the main default comparison, shared compiled operators use
configurations selected during Native calibration and frozen before
measurement. Both systems reuse these selections; checks cover launch
configurations and executable GPU instructions. Native GEMM and default
arithmetic remain enabled. Different batch shapes can still select
different native matrix kernels, and outputs need not match.

The seven levels in Figure~\ref{fig:cumulative-ladder} use
default numerics, native GEMM, and standard compiler autotuning. Batch
invariance is disabled. Each level is measured on the same 256-question
replenished workload, with three ordinary runs and two independent
Nsight Systems 2025.3.1 profiles per answer condition. All 28 profiles
use the same collector and pass checks against the recorded model forward
counts, projection counts, and capture boundaries.

In the held-out evaluation (Table~\ref{tab:heldout}), Native and
\workname{} use vLLM's original linear kernels in both numerical settings.
For the batch-invariant comparison, both systems enable deterministic
reductions in Inductor and disable in-place buffer reuse and combo-kernel
benchmarking.

\end{document}